\documentclass[preprint,12pt]{elsarticle}

\usepackage{amssymb}
\usepackage{amsmath}
\usepackage{lineno,hyperref}
\usepackage{bm}
\usepackage{amsfonts,amssymb,amsmath}
\usepackage{multirow}
\usepackage{array}
\usepackage{color}
\usepackage{tabularx}
\modulolinenumbers[5]

\usepackage{algorithmic}
\usepackage{algorithm}
\usepackage{array}
\usepackage{textcomp}
\usepackage{stfloats}
\usepackage{url}
\usepackage{verbatim}
\usepackage{graphicx}
\usepackage{multirow}
\usepackage{array}
\usepackage{booktabs}
\usepackage{makecell}
\usepackage{colortbl}

\journal{Journal of Pattern Recognition}

\begin{document}

\begin{frontmatter}



\title{Skeleton and Font Generation Network for Zero-shot Chinese Character Generation}

\author[label1]{Mobai Xue}
\author[label1]{Jun Du}
\author[label1]{Zhenrong Zhang}
\author[label1]{Jiefeng Ma}
\author[label1]{Qikai Chang}
\author[label1]{Pengfei Hu}
\author[label2]{Jianshu Zhang}
\author[label2]{Yu Hu}

\affiliation[label1]{organization={NERC-SLIP, University of Science and Technology of China},
            addressline={No.96, JinZhai Road Baohe District},
            city={Hefei},
            postcode={230026},
            state={Anhui},
            country={Chine}
            }
\affiliation[label2]{organization={iFLYTEK Research},
            addressline={No.666 Wangjiang West Road},
            city={Hefei},
            postcode={230088},
            state={Anhui},
            country={China}
            }



\begin{abstract}
Automatic font generation remains a challenging research issue, primarily due to the vast number of Chinese characters, each with unique and intricate structures. Our investigation of previous studies reveals inherent bias capable of causing structural changes in characters. Specifically, when generating a Chinese character similar to, but different from, those in the training samples, the bias is prone to either correcting or ignoring these subtle variations. To address this concern, we propose a novel Skeleton and Font Generation Network (SFGN) to achieve a more robust Chinese character font generation. Our approach includes a skeleton builder and font generator. The skeleton builder synthesizes content features using low-resource text input, enabling our technique to realize font generation independently of content image inputs. Unlike previous font generation methods that treat font style as a global embedding, we introduce a font generator to align content and style features on the radical level, which is a brand-new perspective for font generation. Except for common characters, we also conduct experiments on misspelled characters, a substantial portion of which slightly differs from the common ones. Our approach visually demonstrates the efficacy of generated images and outperforms current state-of-the-art font generation methods. Moreover, we believe that misspelled character generation have significant pedagogical implications and verify such supposition through experiments. We used generated misspelled characters as data augmentation in Chinese character error correction tasks, simulating the scenario where students learn handwritten Chinese characters with the help of misspelled characters. The significantly improved performance of error correction tasks demonstrates the effectiveness of our proposed approach and the value of misspelled character generation.
\end{abstract}





\end{frontmatter}



\section{Introduction}
The field of font generation has blossomed in recent years and has found wide-ranging applications in font design, advertising production, brand design, and beyond. As researchers increasingly focus their attention on this exciting field, novel and more efficient algorithms have been introduced. Nevertheless, the generation of Chinese character fonts continues to pose a significant challenge, due to their intricate structure and vast number of characters. Unlike the font generation challenges faced in other languages, the training sets required to adequately cover all categories of Chinese characters are exceedingly difficult to obtain, and unseen characters may appear during the testing.\par

In recent years, researchers have made significant strides in the field of font generation, achieving notable successes in a number of endeavors. For instance, RD-GAN \cite{RDGAN} can generate unseen glyph in a one-shot setup by introducing a radical extraction module. DM-Font \cite{DMfont} and LF-Font \cite{LFfont} improve the generative quality by learning component-wise style representation. CG-GAN \cite{CGGAN} supervises the generator to decouple content and style at a more granular level. Despite these advances, we observe that the existing methods are unable to notice tiny differences when generating characters, whose glyphs are much similar to those in the training set. In the testing stage, unseen categories degrade into the similar and seen ones, which seems to be the result of biases intrinsic to present font generation models.\par

\begin{figure}[!t]
	\centering
	\includegraphics[width=0.6\linewidth]{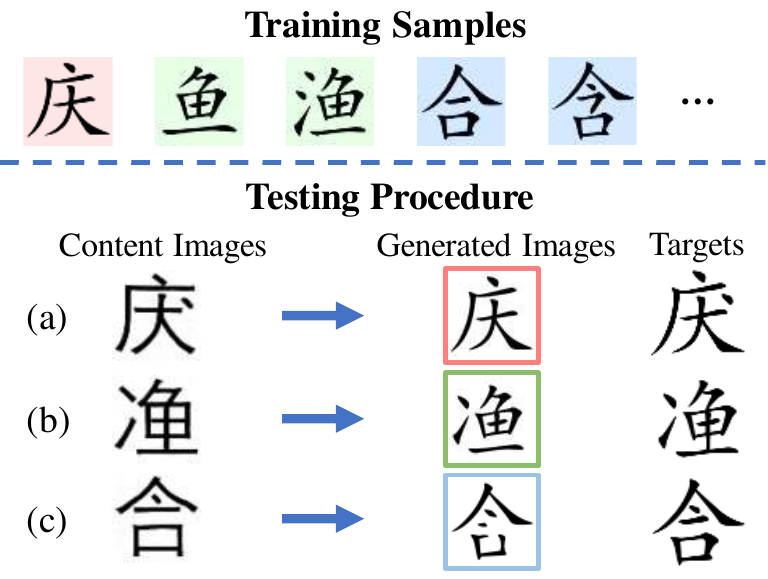}
	\caption{The examples of error-generated results produced by CG-GAN: (a) the failure to capture an additional point in a content image, (b) the retention of the left component but replacing the right to a high-frequency radical, and (c) the inability to model components that bear similarity to multiple training samples.}
	\label{fig.ex_error}
\end{figure}

In Figure \ref{fig.ex_error}, several instances of error-generated results produced by font generation method are illustrated, with CG-GAN chosen as the exemplary method. Figure \ref{fig.ex_error}(a) presents a content image similar to the red training sample. However, the model fails to pay attention to the additional component within the content image, resulting in an output image that is virtually indistinguishable from the training sample. In Figure \ref{fig.ex_error}(b), the right component of content image is similar to the ones in the green training samples, which frequently appears in training set. Although the model succeeds in preserving the left component, the right component is nevertheless corrected to the high-frequency component. In Figure \ref{fig.ex_error}(c), the content image appears to be situated midway between the two blue training samples and the distortions appear in the place where the content image is conflict with the training samples. The failure of modeling components may be caused by the conflict between the model memory and the glyph of content image. Two potential reasons for the occurrence of biases in the model may be inferred: 1) Due to the small number of categories used for training, model learns only a narrow range of combinations of components. The content features are likely to be bias and divide the feature space to limited sub-spaces. 2) The style features are integrated with content features in simple ways, such as addition or concatenation. The information of components' glyph and spatial location in style images is largely ignored, which possibly leads to the form and font bias of components.\par

In this paper, we present a Skeleton and Font Generation Network (SFGN) to achieve more robust generation of Chinese characters. The aim of the proposed approach is to alleviate the detrimental effects of bias in both content and style features, which is achieved through dividing the generation task into two sub-tasks: glyph generation and font generation. We believe that beginning with sub-tasks in lower difficulty levels is more conducive to the model learning. Our approach consists of a skeleton builder and font generator and divides the generation process into two stages. Firstly, the skeleton builder utilizes component-level captions to construct the glyph of a Chinese character and output a content feature. Unlike previous font generation methods that extract features from image inputs, the skeleton builder operates without the involvement of content images. Caption inputs make content features more flexible and enable the model to learn more character categories. The skeleton builder possesses the capability of zero-shot learning and can generate unseen glyphs. Secondly, the font generator extracts style features from style images and integrates the content and style features to produce a Chinese character. Current font generation methods tend to simply integrate content and style features through addition or concatenation, ignoring the glyph and spatial location of components in style features. Thus, we introduce a transitive-attention mechanism to learn the alignments of components among content and style features. The font generation effectively transfers components from style features to content features, modeling font style at a more instructive and granular level.\par

The proposed SFGN is inspired by two human behaviors \cite{shen2005investigation}: 1) people can write an unknown character through a description of the known basic components and structural relationships, and 2) when imitating other fonts, people typically copy components from existing characters in target font and adaptively adjust their spatial relationship. Such human learning schemes are adopted in our proposed skeleton builder and font generator respectively. Specifically, the skeleton builder can create unseen categories based on the learned components, achieving zero-shot glyph generation. The font generation achieves the function of copying components from style features to content features.\par

Extensive experimental results demonstrate the superiority of our approach. Firstly, we compare our proposed SFGN with state-of-the-art methods on glyph and font generation tasks. We also discuss the performance on misspelled characters which contain a large number of samples that differ slightly from the right ones. Our method outperforms other methods and reduce the adverse impact of biases on glyph. Furthermore, we believe that misspelled characters play a significant role in Chinese character teaching \cite{lam2011chinese}. Therefore, we design a series of augmentation experiments on handwritten Chinese character error correction tasks to simulate the impact of misspelled characters for the students to learn Chinese characters. In these experiments, the images generated by font generation methods are employed as augmented data. As expected, the performance of error correction tasks improves significantly, which proves the wide applicability and practical value of misspelled character generation in education.\par

The main contributions of our approach are as follows:
\begin{itemize}
\item We propose a Skeleton and Font Generation Network (SFGN) to mitigate the negative impact of bias and achieve low-resource font generation. 
\item A novel transitive-attention mechanism is introduced to learn the alignment of components between content and style features. 
\item Our proposed model demonstrates superior image generation performance in comparison to state-of-the-art glyph and font generation methods, particularly with regards to the generation of misspelled characters. The improved performance in augmentation experiments on error correction task highlights the educational value of misspelled character generation.

\end{itemize}

\section{Related Work}
\subsection{Stroke and Radical in Chinese Character}
\begin{figure}[!t]
	\centering
	\includegraphics[width=0.6\linewidth]{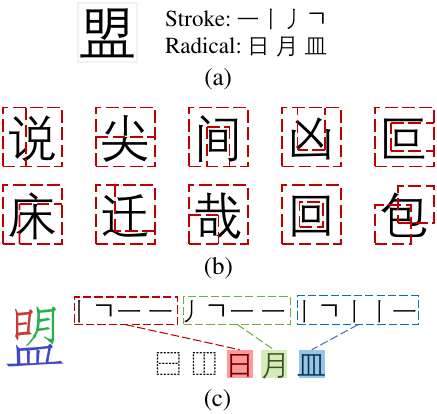}
	\caption{The examples of components and structures in Chinese characters. (a) shows the strokes and radicals in the example character, (b) displays 10 structures between radicals and (c) shows the captions in radical and stroke level.}
	\label{fig.background}
\end{figure}

Chinese characters are hieroglyph with a large number of existing categories. Due to the complex 2D structure of each character, modeling Chinese characters is challenging and related tasks are more difficult. However, there is inherent law for the structure of Chinese characters, as each Chinese character is composed of 8 types of strokes, i.e. the smallest basic components. The high-frequency stroke combinations are defined as radicals, and 500 radicals are adequate to describe more than 20,000 Chinese characters \cite{CH_back_2}. Figure \ref{fig.background}(a) shows the strokes and radicals in the example character and (b) displays 10 structures of radicals. All Chinese characters can be decomposed into radicals and their spatial relationships as captions. Radical-level caption can uniquely determine the structure for each character. Figure \ref{fig.background}(c) shows the captions in radical and stroke level. Nevertheless, due to the few quantities of strokes and undefinable relationship among strokes, the captions in stroke-level cannot denote each Chinese character uniquely and are not usually used directly.

\subsection{Zero-shot Glyph Generation}

Zero-shot glyph generation aims to generate unseen Chinese characters that are not collected in training set. As the performance of radical-based recognition methods is mainly influenced by the amount of character categories in the training set, glyph generation is proposed to provide more radical combinations and improve the recognition accuracy. RCN \cite{RTN-G} proposes a tree-structured encoder based on recursive algorithm, which can create unseen characters based on customized inputs. Later, RTN-G \cite{RCN} proposes a transformer-based generation network that decrease computational complexity and improves image quality. Nevertheless, neither of them can modify stroke-level details, which is one main limitation for the construction of higher-quality content features. On the basis of above works, we model the glyph of Chinese characters through both radical and stroke level captions. Our proposed skeleton builder also introduces a scale correction mechanism to integrate the radical and stroke in different scales, which outperforms RCN and RTN-G significantly.

\subsection{Font Generation}
Font generation can be categorized into single font generation and arbitrary font generation. Single font generation methods \cite{strokegan, SGCE, zigan} solely focus on the transformation between two designated font styles, necessitating model retraining whenever a new font is introduced. Arbitrary font generation aims to create a complete font library in the required style given only a few reference images. Several methods \cite{several_I2I_1, several_I2I_2} regard it as an image-to-image translation problem as both tasks learn a mapping from the source domain to the target domain. For instance, ``zi2zi'' \cite{zi2zi} learns multiple font styles by adding a pre-defined style category embedding but does not support the generation of unseen font styles. DC-font \cite{DCfont} learns the transformation relationship between two fonts in deep space via the feature reconstruction network. Since ``zi2zi'' and DC-font construct the mapping relationship between seen font styles only, all of them cannot generalize to unseen styles. After that, EMD \cite{EMD} and SA-VAE \cite{SA-VAE} disentangle the font style and content representation and can generate unseen font styles. However, they model fonts in character-level and fail to capture local style patterns \cite{TMM-font-1}. Later, some component-based methods \cite{ccfont} are proposed. For example, LF-Font \cite{LFfont} and DM-Font \cite{DMfont} improve the generation quality by learning component-wise style representation. Specifically, LF-Font can be extended to unseen styles conditioned on the component-wise style features. DM-Font employs a dual-memory architecture for font generation, which requires a reference set containing all the components to extract the stored information. Furthermore, to tackle the requirement of large amounts of paired data for pixel-level strong supervision, DG-Font \cite{DGFont} achieves unsupervised learning by introducing a deformation skip connection. MX-Font \cite{MXFont} employs a multi-headed encoder to extract different localized features in a weak component supervised manner. CG-GAN \cite{CGGAN} supervises the generator at the component level for both styles and contents which can generalize to cross-lingual font generation. Nevertheless, we find that all of the above works do not fully utilize the vital information of component glyph and spatial location in style images \cite{TMM-font-2}. Despite FsFont \cite{FsFont} can learn the spatial correspondence between content and style images, it fails to capture spatial correspondence in complex handwritten scene. To this end, we propose a font generator to adaptively extract and apply the components in style images with different categories and fonts. Our approach can draw a Chinese character through copying components from style images, which has impressive performance and good model interpretability.

\subsection{Chinese Character Error Correction}
Handwritten Chinese Character Error Correction (HCCEC) is a novel task, which aims to assess the correctness of handwritten characters and correct potential misspelled characters. TAN \cite{TAN} first presents a novel tree-structured analysis network for HCCEC, which employs a tree decoder and successfully achieves the error correction of handwritten Chinese characters. Notably, HCCEC is easily confused with Chinese Grammatical Error Correction (CGEC)\cite{CGEC1}. CGEC focuses on detecting and correcting grammatical errors in text\cite{CGEC2}, while HCCEC judges the correctness of single handwriting character and predicts the expected word that the author intends to write. \par

Moreover, HCCEC is a variation of offline handwritten Chinese character recognition (HCCR). HCCR can be categorized into character-based \cite{TMM-recognition-1, TMM-recognition-2} and radical-based \cite{rbased-r-1, rbased-r-3} methods. Although some radical-based methods are proposed for HCCR, they can also be applied to HCCEC. For instance, RAN \cite{RAN} is proposed to decompose a Chinese character into a sequence of radicals and structures. HDE \cite{HDE} proposes a hierarchical decomposition embedding method to represent a Chinese character with a semantic vector. In our approach, to prove the effectiveness and value of generated misspelled characters on Chinese characters teaching, we employ HCCEC methods to simulate the scene that students learn Chinese characters.


\begin{figure*}[!t]
	\centering
	\includegraphics[width=1.0\textwidth]{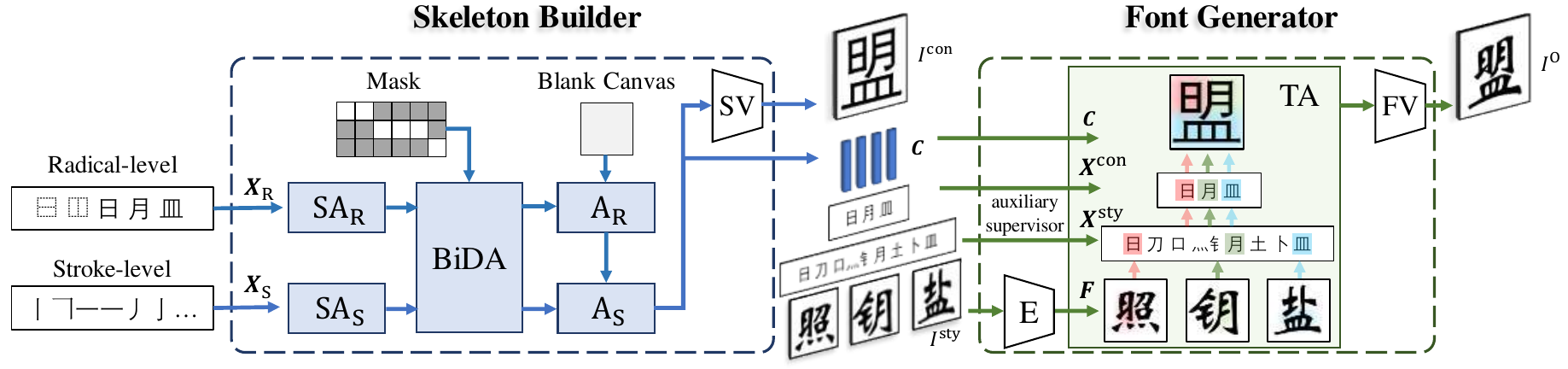}
	\caption{Overview of the proposed method SFGN, which contains a skeleton builder and a font generator. Firstly, the skeleton builder creates character images in standard font according to input radical and stroke level captions. Then, the font generator transfers the character images from standard font to target fonts, whose glyphs are determined by content features, while the style of the fonts is characterized by style images. In skeleton builder, ``A'' denotes the fundamental attention block, ``SA'' denotes self-attention block and ``BiDA'' denotes bi-directional attention block, where the subscripts ``R'' and ``S'' represent radical and stroke respectively. In the font generator, ``TA'' denotes the proposed transitive-attention block. ``SV'' and ``FV'' are visualization render modules to map features to images.}
	\label{fig.archi}
\end{figure*}

\section{Methodology}
Our proposed SFGN architecture comprises a skeleton builder and a font generator, as depicted in Figure \ref{fig.archi}. This research aims to generate Chinese character images, whose glyphs are defined by input captions, and font styles maintain consistency with the input style images. The skeleton builder takes captions at the radical and stroke levels as input and generates content features that contain glyph information, as introduced in Section \ref{sec.SB}. The font generator first extracts style features from style images and then integrates the content and style features to synthesize character images, as illustrated in Section \ref{sec.FG}. Additionally, the radical-level captions of both the skeleton and style images are utilized for auxiliary supervision. Multiple loss functions are employed to optimize the model and effectively utilize labeled and unlabeled data, which are explicated in Section \ref{sec.loss}. By leveraging these modules, our SFGN achieves superior performance in generating high-quality Chinese characters that accurately fit the descriptions of input captions.

\subsection{Skeleton Builder}\label{sec.SB}
Exhibited in the left-side segment of Figure \ref{fig.archi}, the skeleton builder aims to transfer input radical and stroke level captions to corresponding character images in standard font through multiple attention mechanism blocks and a skeleton render module. Specifically, the fundamental attention \cite{attn} blocks are denoted by ``A'', while self-attention \cite{Transformer} blocks are represented by ``SA'', and bi-directional attention \cite{BiDA} blocks are designated as ``BiDA''.\par

For clarity, we first introduce the general kernel function $\alpha$ of these attention blocks, followed by a brief description of the different attention mechanisms within $\alpha$. Typically, an attention function maps a query and a set of key-value pairs to an output where all four elements are vectors. An attention weight is calculated by a scaled dot-product function, which can be expressed as follows:
\begin{equation}\label{eq.attn}
    \begin{aligned}
      \text{attn}(\boldsymbol{Q},\boldsymbol{K}) &= \text{SoftMax}\left( \frac{\boldsymbol{QK}^\top}{\sqrt{D}} \right)
    \end{aligned}
\end{equation}
where $D$ is the dimension of the model and $\boldsymbol{Q},\boldsymbol{K}$ are $D$-dimensional vectors. In our approach, we employ multi-head attention mechanism as the kernel function $\alpha$.
\begin{equation}
    \begin{aligned}
      \alpha(\boldsymbol{Q},\boldsymbol{K},\boldsymbol{V}) &= \left[ \boldsymbol{h}_1,\boldsymbol{h}_2,\cdots,\boldsymbol{h}_M \right]\boldsymbol{W}^\text{O}\\ 
      \text{where}\quad \boldsymbol{h}_j &= \text{attn}(\boldsymbol{Q}\boldsymbol{W}^\text{Q}_j, \boldsymbol{K}\boldsymbol{W}^\text{K}_j)\boldsymbol{V}\boldsymbol{W}^\text{V}_j
    \end{aligned}
\end{equation}
where $M$ is the number of heads, $[\cdots]$ represents the concatenation operation, $\boldsymbol{W}^\text{Q}_j,\boldsymbol{W}^\text{K}_j,\boldsymbol{W}^\text{V}_j\in\mathbb{R}^{D\times D}$ are the projection matrices of $j^\text{th}$ head and $\boldsymbol{W}^\text{O}\in\mathbb{R}^{DM\times D}$.\par

Given captions in radical and stroke level, we first map the sequences of symbol representations to the sequences of continuous representations and obtain $\boldsymbol{X}_\text{R}$ and $\boldsymbol{X}_\text{S}$. Position embedding $\boldsymbol{P}$ \cite{PE} is employed here to interpret the input by distinguishing the relative positional relationships between the radical/stroke symbols.

\begin{equation}\label{eq.cap_emb}
  \boldsymbol{X} = \{\boldsymbol{x}_1+\boldsymbol{P}_1,\cdots,\boldsymbol{x}_L+\boldsymbol{P}_L\},\quad \boldsymbol{x}_i, \boldsymbol{P}_i\in\mathbb{R}^D
\end{equation}
where $L$ is the length of a caption and the positional embedding PE is defined as:
\begin{align}
        \boldsymbol{P}_{i,2d} &=\sin\left(\frac{i}{10000^{2d/D}}\right)\\
        \boldsymbol{P}_{i,2d+1} &= \cos\left(\frac{i}{10000^{2d/D}}\right)
\end{align}
where $i$ is the position and $d$ is the dimension. The representations $\boldsymbol{X}_\text{R}$ and $\boldsymbol{X}_\text{S}$ are fed into self attention blocks, which enable the exchange of internal information within the radical and stroke level representations. Self-attention blocks allow every position to attend to all positions, where keys, values, and queries are derived from the same source.
\begin{equation}
  \boldsymbol{X}^\prime=\text{SA}(\boldsymbol{X}) = \alpha(\boldsymbol{X}, \boldsymbol{X},\boldsymbol{X})
\end{equation}

Then, the bi-directional attention block integrates the information from radical and stroke representations in both directions. When both the query and key contain essential information, this block comes into full operation. The correspondence between strokes and radicals is recorded as a mask, which is subsequently utilized during bi-directional attention calculation. The bi-directional attention mechanism can be articulated as follows: 
\begin{equation}\label{eq.bida}
  \text{BiDA}(\boldsymbol{X}_\text{R}^\prime,\boldsymbol{X}_\text{S}^\prime)=\left\{
  \begin{aligned}
    \boldsymbol{X}_\text{R}^{\prime\prime}: & \enspace \alpha(\boldsymbol{X}_\text{R}^\prime , \boldsymbol{X}_\text{S}^\prime ,\boldsymbol{X}_\text{S}^\prime ) \\
    \boldsymbol{X}_\text{S}^{\prime\prime}: & \enspace \alpha(\boldsymbol{X}_\text{S}^\prime , \boldsymbol{X}_\text{R}^\prime ,\boldsymbol{X}_\text{R}^\prime )
  \end{aligned}
  \right.
\end{equation}
where $\boldsymbol{X}^{\prime\prime}\in\mathbb{R}^{L\times D}$. The subscript notation ``R'' and ``S'' correspond to radical and stroke respectively. According to the previous work RTN-G\cite{RTN-G}, the representations of radicals $\boldsymbol{X}_\text{R}^\prime$ are prone to being orthogonal, while the representations of strokes $\boldsymbol{X}_\text{S}^\prime$ are often highly alike due to their limited categories. Accordingly, through secondary modeling using the bi-directional attention block, the representations of similar radicals are closer. Meanwhile, stroke representations that belong to different radicals become more distinguishable.\par

To construct a 2-dimensional content features through those 1-dimensional representations, we employ a series of vectors initialized with position embedding of size $4\times 4\times D$, called ``blank canvas'' $\boldsymbol{B}$.
\begin{equation}
    \boldsymbol{B} = \{\boldsymbol{P}_{1}, \boldsymbol{P}_{2},\cdots, \boldsymbol{P}_{16}\}, \quad\boldsymbol{P}_{i}\in\mathbb{R}^D
\end{equation}

A fundamental attention block $\text{A}_\text{R}$ fills the radical representations into appropriate location and generate the vectors $\boldsymbol{B}^\prime$ containing radical-level information.
\begin{equation}
    \boldsymbol{B}^\prime = \text{A}_\text{R}(\boldsymbol{X}_\text{R}^{\prime\prime}, \boldsymbol{B})=\alpha(\boldsymbol{X}_\text{R}^{\prime\prime} , \boldsymbol{B} ,\boldsymbol{B} )
\end{equation}
 where $\boldsymbol{B}^\prime \in \mathbb{R}^{4\times 4\times D}$. Notably, each $D$-dimensional vector in the ``blank canvas'' representation corresponds to a $1/16$ proportion of the total image area, as all radicals are at such scale. Since strokes are typically smaller in size than radicals, treating them at an equal scale could cause distortion, which is proven in the ablation study in Section \ref{sec.ex_ab}. Consequently, we upsample $\boldsymbol{B}^\prime$ and double its size, resulting in $\boldsymbol{B}^{\prime\prime} \in \mathbb{R}^{8\times 8\times D}$. Furthermore, another fundamental attention block $\text{A}_\text{S}$ is employed to supplement the canvas with stroke-level information.

\begin{equation}
  \boldsymbol{B}^{\prime\prime\prime} = \text{A}_\text{S}(\boldsymbol{X}_\text{S}^{\prime\prime}, \boldsymbol{B}^{\prime\prime}) = \alpha(\boldsymbol{X}_\text{S}^{\prime\prime} , \boldsymbol{B}^{\prime\prime} ,\boldsymbol{B}^{\prime\prime} )
\end{equation}\par

Lastly, we construct a content features $\boldsymbol{C}\in\mathbb{R}^{4\times 4\times D}$, which is supervised at the radical level in the font generator. To reduce the area of $\boldsymbol{B}^{\prime\prime\prime}$ and output the content features, a convolution \cite{cnn} followed by $2\times2$ average pooling is employed. Furthermore, a deconvolution-based \cite{deconv} skeleton render module ``SV'' transforms the canvas $\boldsymbol{B}^{\prime\prime\prime}$ into a content image $\hat{I}^\text{con}$ in standard font. The quality of content images is related to the quality of content features.

\subsection{Font Generator}\label{sec.FG}
Illustrated in the right-hand portion of Figure \ref{fig.archi}, the font generator aims to transfer character images from standard font to target fonts, where the glyphs are determined by input content features and font styles are characterized by input style images. Given content features, style images and the captions of them, the font generator creates images of Chinese characters with corresponding content and font styles. \par

Instinctively, when attempting to write characters in a new font, it is more natural for humans to copy components from existing characters in the desired font. Typically, people will first search for required components, then place them in the appropriate location. However, current methods approach style features in a fixed way, regardless of their origin from varying character categories. They integrate content and style features through simplistic operations, such as addition or concatenation, while also disregarding critical information pertaining to component glyph and spatial location. Therefore, we introduce a transitive-attention mechanism in our font generator, enabling our model to learn alignments between content and style features at radical level.\par

As shown in the right half of Figure \ref{fig.archi}, our proposed font generator consists of a DenseNet-based \cite{densenet} style encoder ``E'', a transitive-attention block ``TA'' and a font render module ``FV''. Firstly, to achieve the goal of copying components from style images, we use $T$ style images. The style images need to provide all the required radicals of the content images. Based on the statistics of the adopted Chinese characters, we set $T$ as 5. These style images are concatenated together and encoded by the style encoder into style features $\boldsymbol{F}\in \mathbb{R}^{4\times 4\times D}$. Given content features $\boldsymbol{C}$, style features $\boldsymbol{F}$, the font generator should adaptively capture and send the required components from style features to content features.\par


Consequently, we employ the radical-level captions of content and style images to build bridges between content and style features. While the alignment at the stroke level is challenging, the stroke-level captions are not employed here. The captions of content images are mapped to continuous sequences $\boldsymbol{X}_\text{R}^\text{con}$, while those of style images are mapped to $\boldsymbol{X}_\text{R}^\text{sty}$. Notably, they utilize the same embedding weights as those belonging to the skeleton builder as mentioned in Equation \ref{eq.cap_emb} and positional embedding are also used. We expect the font generator can learn the alignment in a chain, from style images, to captions, and returning to content images.\par

Nevertheless, current attention mechanisms only support 2 inputs and cannot produce the attention among a series of query-key pairs. Accordingly, we introduce a transitive-attention mechanism, whose kernel function is denoted as $\beta$.
\begin{align}
    \hat{\beta}(\boldsymbol{Q}_1,\cdots,& \boldsymbol{Q}_{N-1}, \boldsymbol{K}_2, \cdots ,\boldsymbol{K}_N), \notag\\
    \label{eq.beta} &=\prod_{i=1}^{N-1}\exp(\boldsymbol{Q}_i)\exp( \boldsymbol{K}_{i+1}^\top)
\end{align}
where $N$ denotes the number of query-key pairs. Each product term in Equation \ref{eq.beta} represents the spatial attention weight from the $(i+1)^\text{th}$ term to the $i^\text{th}$ one, and the exponent function ensures their non-negativity. Moreover, we normalize $\hat{\beta}$ to modify its variance, then obtain $\beta$ through SoftMax. Each query and key are supposed to follow normal distributions $\mathcal{N}(0,1)$ and $\beta$ can be expressed as:
\begin{equation}
    \beta =\text{SoftMax}\left(\frac{\hat{\beta}}{\sqrt{V_{2N-2}}}\right)
\end{equation}
where the variance $V_n$ depending on the number $n$ of used queries and keys is denoted as:
\begin{align} \label{eq.vn}
    V_n =D^{n-1} \left[ \sum_{i=0}^{n} (D-1)^i C_n^i \cdot e^{2n-i} \right] - e^{n}D^{2(n-1)} 
\end{align}
where combination number $C_N^i=N!/((N-i)!i!)$ and $N=4$. Based on $\beta$, we define the transitive-attention mechanism as follows:
\begin{equation}
    \begin{aligned}
        \text{TA}(\boldsymbol{X}_1, \cdots, \boldsymbol{X}_N) &= \beta (\boldsymbol{X}_1\boldsymbol{U}^\text{Q}_1, \cdots,\boldsymbol{X}_{N-1}\boldsymbol{U}^\text{Q}_{N-1}\\
        &\boldsymbol{X}_2\boldsymbol{U}^\text{K}_{2}, \cdots,\boldsymbol{X}_{N}\boldsymbol{U}^\text{K}_{N})\boldsymbol{X}_{N}\boldsymbol{U}^\text{V}_{N}
    \end{aligned}
\end{equation}
where $\boldsymbol{U}^\text{Q}_j, \boldsymbol{U}^\text{K}_j, \boldsymbol{U}^\text{V}_j \in \mathbb{R}^{D\times D}$ are the projection matrices for $j^\text{th}$ query, key and value. The transitive-attention block builds character representations $\boldsymbol{X}_\text{C}$ through content features $\boldsymbol{C}$, style features $\boldsymbol{F}$, content representations $\boldsymbol{X}_\text{R}^\text{con}$ and style representations $\boldsymbol{X}_\text{R}^\text{sty}$.
\begin{equation}
	\boldsymbol{X}_\text{C}  =\text{TA}(\boldsymbol{C},\boldsymbol{X}_\text{R}^\text{con} ,\boldsymbol{X}_\text{R}^\text{sty} , \boldsymbol{F})
\end{equation}
where $\boldsymbol{X}_\text{C}\in\mathbb{R}^{4\times 4 \times D}$.\par

Additionally, we apply a mask to split and align the style captions and images in the transitive-attention block. Finally, the character representations $\boldsymbol{X}_\text{C}$ are processed by the deconvolution-based font render module to obtain the final output $\hat{I}^\text{gen}$. $\hat{I}^\text{gen}$ retains the glyph of $I^\text{con}$ and presents the font style of $I^\text{sty}$.\par

Moreover, we delve deeper into the transitive-attention mechanism. It constructs a transmission path of attention, manifested through three distinct attention maps: 1) the spatial relationship $a_\text{I2T}$ between style features and their captions (from $\boldsymbol{F}$ to $\boldsymbol{X}_\text{R}^\text{sty}$), 2) the similarity $a_\text{T2T}$ among the radicals of captions (from $\boldsymbol{X}_\text{R}^\text{sty}$ to $\boldsymbol{X}_\text{R}^\text{con}$), and 3) the spatial relationship $a_\text{T2I}$ between content features and their captions (from $\boldsymbol{X}_\text{R}^\text{con}$ to $\boldsymbol{C}$). Interestingly, these three attention maps correspond to human behaviors when imitating fonts: 1) identifying the components in the target font style, 2) analyzing which components are required, and 3) transferring the required components from style images to the appropriate locations. Furthermore, we can also re-write the Equation \ref{eq.beta} as follows:
\begin{equation}
	\label{eq.re-beta} \hat{\beta} = \exp(\boldsymbol{Q}_1)\left[ \prod_{i=2}^{N-1}\exp( \boldsymbol{K}_{i}^\top)\exp(\boldsymbol{Q}_i) \right] \exp(\boldsymbol{K}_N^\top)
\end{equation}

As mentioned above, the objective of $\beta$ is to establish radical-level alignment by matching $\boldsymbol{Q}_1$ and $\boldsymbol{K}_N$. Equation \ref{eq.re-beta} provides a representation of $\beta$ as the product of $\boldsymbol{Q}_1$ and $\boldsymbol{K}_N$, augmented by a succession of corrective terms. Every term may be construed as representing a channel attention weight generated from the $i^\text{th}$ input. These attention weights regulate the channel distribution, thereby enabling $\boldsymbol{Q}_1$ to become more adaptable to match $\boldsymbol{K}_N$. Notably, the $i^\text{th}$ channel attention weight is entirely reliant on the $i^\text{th}$ input. In our approach, the $2^\text{nd}$ to $(N-1)^\text{th}$ inputs are the caption representations, which implies the channel attention learns the glyph distribution through caption inputs. It works simultaneously to amplify the channels that contain crucial radical information and ensures that the content features match with accurate vectors present in style features.

\subsection{Objectives}\label{sec.loss}
During the training, we utilize several loss functions to optimize our model with both labeled and unlabeled data. As the unavailability of misspelled character images and incomplete character categories in certain fonts, our models apply varying techniques to available and unavailable images. To differentiate between the two, we define the set of available images as $\mathbb{U}^\text{r}$ and the set of unavailable images as $\mathbb{U}^\text{f}$.

\textbf{Pixel Loss} We constrain the generated images $\hat{I}^\text{con}$ and $\hat{I}^\text{gen}$ in pixel-level by pixel loss $\mathcal{L}_p$, which uses root mean square error (RMSE) to measure the distortion among generated images and target images. Notably, $\mathcal{L}_p$ is applied in both skeleton builder and font generator, and only used for available samples.

\begin{equation}
	\mathcal{L}_p=\mathop{\mathbb{E}}\limits_{I\in \mathbb{U}^\text{r}}\sqrt{\frac{1}{HW}\sum_i^H\sum_j^W(\hat{I}_{i,j} - I_{i,j})^2}
\end{equation}
where $H,W$ are the height and width of images.\par

\begin{figure}[!t]
	\centering
	\includegraphics[width=0.7\linewidth]{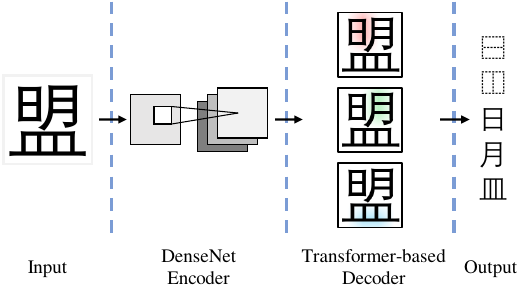}
	\caption{The framework of RTN-R, specifically the adopted radical-based recognition method in content loss.}
	\label{fig.rtn}
\end{figure}

\textbf{Content Loss}
In order to ensure the accuracy of both the components and structures in generated images, we introduce a content loss function $\mathcal{L}_c$. A radical-based recognition model $\mathcal{R}$ is employed to predict the captions of input images, where we employ RTN-R \cite{RTN-R} in our approach. This recognition model consists of a DenseNet encoder and a Transformer-based decoder as visualized in Figure \ref{fig.rtn}. The encoder extracts high-dimensional representations from images, while the decoder predicts symbols step-by-step. Cross entropy \cite{CE} is employed to measure the distance between image captions $y$ and the predictions, while $\mathcal{L}_c$ can be represented as:
\begin{equation}
    \begin{aligned}
	\mathcal{L}_c &= \mathop{\mathbb{E}}\limits_{I\in \mathbb{U}^\text{r}}\left[ -\frac{1}{L}\sum_l^L y_l \log \mathcal{R}(I)_l \right]\\
        & + \mathop{\mathbb{E}}\limits_{I\in \mathbb{U}^\text{f}}\left[ -\frac{1}{L}\sum_l^L y_l \log \mathcal{R}(\hat{I})_l \right]
    \end{aligned}
\end{equation}
where $L$ is the length of captions. The recognition model is optimized on available samples and frozen on unavailable samples, thus it can guide our method to synthesize images with correct contents. $\mathcal{L}_c$ is applied in both skeleton builder and font generator.\par

\textbf{Guided Loss}
Model training based on the transitive-attention mechanism is challenging, since the alignment of components between content and style features is implicit. In order to regulate the transitive-attention, we introduce a guided loss function $\mathcal{L}_g$. As discussed in Section \ref{sec.FG}, the transitive-attention process can be regarded as the multiplication of three individual attention maps, whereby Equation \ref{eq.beta} can be restated as:
\begin{align}
	\hat{\beta}=\hat{a}_\text{I2T}\cdot \hat{a}_\text{T2T}\cdot \hat{a}_\text{T2I}
\end{align}
where ``T'' in subscript represents the texts, ``I'' represents the images and ``T2I'' means the attention weight with texts input as query and images as key-value pair. Furthermore, each individual attention map is denoted as:
\begin{equation}
    a = \text{SoftMax} \left( \frac{\hat{a}}{V_2} \right)
\end{equation}
where $V_2$ is defined by Equation \ref{eq.vn}.\par

Notably, $a_\text{T2I}$ represents the text-to-image relationship, which can be considered as the kernel mapping of the radical-level fundamental attention block in skeleton builder. Moreover, $a_\text{I2T}$ represents the image-to-text relationship, which is the kernel mapping of a recognition task. It's natural to use the attention maps produced by trained generation and recognition models as the guidance for transitive-attention mechanism. When a content image $I^\text{con}$ is used in transitive-attention block, the attention map $a_\text{G}$ in the process of generating $\hat{I}^\text{con}$ in skeleton builder can be the guidance for $a_\text{T2I}$. Similarly, the attention map $a_\text{R}$ in the process of recognizing $I^\text{sty}$ in skeleton builder can be the guidance for $a_\text{I2T}$, when a style image $I^\text{sty}$ is used in transitive-attention block. The guided loss $\mathcal{L}_g$ is defined as follows:
\begin{equation}
    \mathcal{L}_g= \mathop{\mathbb{E}}\limits_{I} \left[ \left\Vert a_\text{T2I} - a_\text{G} \right\Vert_2 + \left\Vert a_\text{I2T} -a_\text{R} \right\Vert_2 \right]
\end{equation}
where, $a_\text{G}$ and $a_\text{R}$ are calculated without gradients. The guided loss is applied on both available and unavailable samples.

\textbf{Full Objective}
The objective $\mathcal{L}_\text{SB}$ of skeleton builder and the objective $\mathcal{L}_\text{FG}$ of font generator can be described as:
\begin{align}
	\label{eq.SB}\mathcal{L}_\text{SB} &= \mathcal{L}_p + \lambda_c\mathcal{L}_c\\
    \label{eq.FG}\mathcal{L}_\text{FG} &= \mathcal{L}_p + \lambda_c\mathcal{L}_c + \lambda_g\mathcal{L}_g
\end{align}
where $\lambda_c$ and $\lambda_g$ are tuned hyper-parameters.

\section{Task and Experiment Setting}

\subsection{Dataset}
\textbf{Standard Font Set.} We denote ``Arial'' font as the standard font. The dictionary contains 27,533 Chinese characters, where 20,000 categories belong to training set and 7,533 categories  belong to testing set.\par
\textbf{Calligraphic Font Set.} Since the methods utilize diverse datasets in their original papers, we collect a dataset containing 200 fonts and utilize unified dataset to ensure the fairness of experiments. The dictionary contains 3,755 commonly used Chinese characters that can be decomposed by 415 radicals. The training set contains 3,000 Chinese characters in 180 fonts. We evaluate the generalization ability on two test sets: One set includes the 150 seen fonts with 755 unseen characters per font. Another is the remaining 20 unseen fonts with 755 unseen characters per font.\par
\textbf{Handwriting Set.} We employ the dataset that introduced by \cite{TAN}, which contains the samples from 5,500 common used character classes and 570 misspelled character classes and the images are clear. The right character classes are split into 5,000 training classes and 500 validation classes. Each character in training set is written by 50 writers (250,000 samples in total), while each character in validation set is written by 200 writers (100,000 samples in total). The testing set contains 11,400 misspelled samples and randomly selected 40,000 right samples written by 20 writers. The testing set is divided into right set and misspelled set and evaluated respectively.\par
\textbf{Korean Set.} We collect a dataset containing 30 Korean fonts to validate the cross-lingual performance of our model. The dictionary contains 400 Korean characters. The training set contains 100 Korean characters in 10 fonts and the testing set contains remaining 300 Korean characters in remaining 20 fonts.\par

\subsection{Task Setting and Evaluation Metrics}
\textbf{Generation Tasks.} The generation task comprises two sub-tasks: glyph generation and font generation. The glyph generation task aims to produce Chinese character images in standard font, whose structure and components match the description of the input captions. The font generation task aims to preserve the structure and components of the content images while changing the font style. \par


The generated images can be divided into two sets: available and unavailable set, depending on whether the target images are collected in dataset. Metrics used for the available set measure the distortion of generated images. Firstly, RMSE (root mean square error) and SSIM \cite{SSIM} (structure similarity index measure) are used to determine whether pixel-level details are retained. Secondly, LPIPS (learned perceptual image patch similarity) \cite{LPIPS} is adopted to quantify the perceptual similarity between the generated and target images. Thirdly, FID (Frechet inception distance score) \cite{FID} is employed to assess the model ability to match the target data domain distribution. In general, higher SSIM, coupled with lower RMSE, LPIPS, and FID, indicate reduced distortion and improved qualities of the generated images. However, the quality of the generated images in the unavailable set can only be determined through visual evaluation.\par


\textbf{Error Correction Task.} There are three sub-tasks in the error correction task: pre-judgement, classification and correction task. The pre-judgement task judges whether the input Chinese character image is written correctly. The classification task aims to accurately categorize the input images into pre-defined groups. The correction task rectifies incorrectly written Chinese characters. Since the prediction task is a binary classification task, we use the $F_1$-score to measure its performance. $F_1$-score reflects the performance of pre-judgement and is calculated by precision and recall.\par


\begin{equation}
	F_1\text{-score} = \frac{2\times \text{precision}\times\text{recall}}{ \text{precision} + \text{recall}}
\end{equation}
Moreover, we use the accuracy rate to assess the classification ability of the error correction model. Ultimately, the correction rate $CR$ can intuitively reflect the error correction ability of the models. Suppose there are $N$ samples in the test set of misspelled characters, $N_c$ represents the samples that are correctly classified and $N_r$ represents the samples whose intended characters are correctly predicted. The correction rate can be expressed as:


\begin{equation}
    CR = \frac{N_c \cap N_r}{N}
\end{equation}

\subsection{Training}
The training model utilized in all experiments has a standardized configuration consisting of the following parameters. All images are resized to $H=64$ and $W=64$. The model dimension $D$ is fixed to 512. We adopt AdaDelta \cite{adadelta} algorithm for model optimization, with the following hyper-parameters: a learning rate of $lrate = 0.1$, a moving average decay factor $\rho$ of 0.95, and a numerical stability constant $\varepsilon$ set to $10^{-4}$. Concerning the hyper-parameters in loss functions, we set $\lambda_c=0.1$ and $\lambda_g=0.1$ in Equations \ref{eq.SB} and \ref{eq.FG}. The experiments are implemented using Pytorch 1.8.0 and an NVIDIA Tesla V100 16G GPU.\par

\textbf{Skeleton Builder.} Each attention block in the skeleton builder comprises four layers, and each layer includes four heads, encompassing attention, self-attention, and bi-directional attention blocks. The skeleton render module consists of 7 deconvolutional layers, where batch normalization \cite{BN} and ReLU \cite{relu} activation functions are employed after each deconvolutional layer. Specifically, there are 3 deconvolutions with a kernel size of $4\times 4$ and stride 2, which are used as expansion layers to double the feature maps, and 4 deconvolutions with a kernel size of $3\times 3$ and stride 1.\par

\begin{figure*}[!t]
	\centering
	\includegraphics[width=1.0\textwidth]{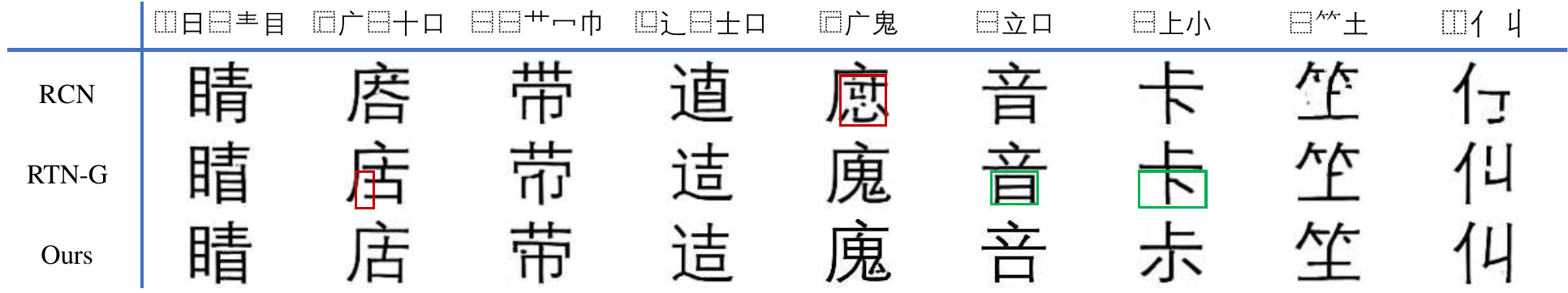}
	\caption{The generated misspelled characters of RCN \cite{RCN}, RTN-G \cite{RTN-G} and our proposed skeleton builder. The radical-level input captions of each generated images are displayed on the top. The green boxes indicate the cases where the generation methods automatically correct misspelled characters to the right ones. The red boxes indicate the cases where the generation methods cannot effectively model a novel combination of character components.}
	\label{fig.SB_show}
\end{figure*}

\textbf{Font Generator.} The number of style image inputs of each character is set to $T=5$, which ensure all required components are included. A DenseNet \cite{densenet} is used as the style encoder, which initially employs a $7\times 7$ convolution layer with 48 output channels before entering the first dense block. Each DenseBlock contains 22 $1\times 1$ convolution layers and 22 $3 \times 3$ convolution layers. After each DenseBlock, we use a transition layer which consists of a $1\times 1$ convolution followed by a $2\times 2$ average pooling to reduce the feature maps by half. The growth rate is set to 24. Batch normalization is used after each convolution layer and the activation function is ReLU.\par

\section{Experiments}\label{sec.exp}
To evaluate the effectiveness of our proposed method, we conducted a series of experiments to compare the performance of the skeleton builder and font generator with state-of-the-art methods. Specifically, in Section \ref{sec.ex_sb}, we compare the performance of the skeleton builder with other glyph generation methods, such as RCN \cite{RCN} and RTN-G \cite{RTN-G}. In Section \ref{sec.ex_fg}, we compare the performance of the font generator with other font generation methods, including single and arbitrary font generation methods. In Section \ref{sec.ex_ab}, we perform ablation experiments to examine the contributions of various modules in our model. Additionally, we believe that misspelled characters play a significant role in teaching Chinese character. To access the practical and educational value of generated misspelled characters, we conduct experiments on error correction tasks in Sections \ref{sec.ex_ec}. We employ the generated images as data augmentation for error correction model TAN \cite{TAN} to imitate the human beings to learn Chinese characters. The evaluation metrics of error correction task can reflect the positive impact of misspelled characters on student learning.

\begin{table}[!t]
	\caption{Performance comparison of RCN \cite{RCN}, RTN-G \cite{RTN-G} and our proposed skeleton builder on glyph generation task.\label{tab.sb}}
	\centering
	\begin{tabular}{c|cccc}
        \hline
		Methods & RMSE $\downarrow$ & SSIM $\uparrow$  & LPIPS $\downarrow$ & FID $\downarrow$\\ 
        \hline
		RCN \cite{RCN} & 0.0211 & 0.7698 & 0.2081 & 41.58\\
		RTN-G \cite{RTN-G} & 0.0176 & 0.8355 & 0.1512 & 6.87\\
		\textbf{Ours} & \textbf{0.0142} & \textbf{0.8914} & \textbf{0.1099} & \textbf{3.44}\\
		\hline
	\end{tabular}
\end{table}

\subsection{Experiments on Glyph Generation}\label{sec.ex_sb}
In this section, we evaluate the performance of skeleton builder on the glyph generation task, which transfers the text captions to Chinese character images with accurate components and structures in standard fonts. The precision of generated images is crucial for the subsequent font generation task, while higher-quality generated results reflect the higher-quality content features. To evaluate the effectiveness of our skeleton builder, we conducted experiments on the standard font dataset.\par

\begin{figure*}[!t]
	\centering
	\includegraphics[width=1.0\textwidth]{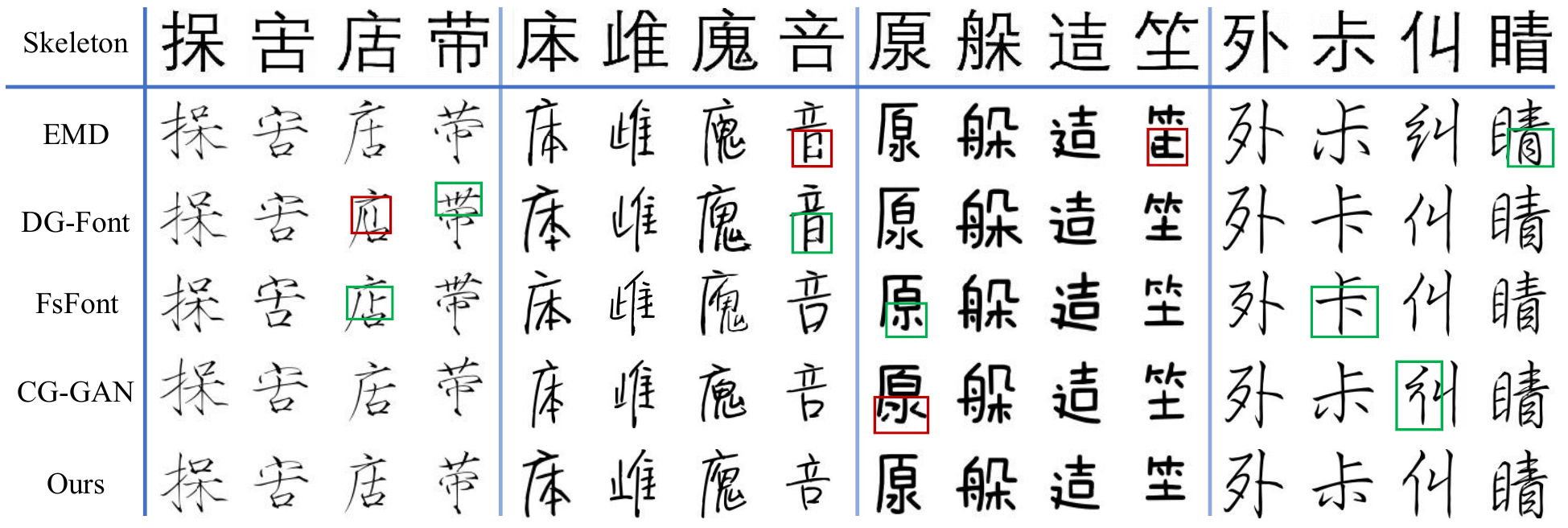}
	\caption{The generated misspelled characters of EMD \cite{EMD}, DG-Font \cite{DGFont}, FsFont \cite{FsFont}, CG-GAN \cite{CGGAN} our proposed font generator. The misspelled content images produced by skeleton builder of each generated images are displayed on the top. The green boxes indicate cases where the font transfer methods automatically correct misspelled characters to the correct ones. The red boxes indicate cases where the font transfer methods cannot effectively capture the glyph of content images.}
	\label{fig.FG_show}
\end{figure*}

Table \ref{tab.sb} demonstrates the generation quality of our method compared with state-of-the-art methods on the test set. Our proposed method achieves the best results for all four indicators, suggesting that our method is capable of generating images with less distortion. In comparison to RCN and RTN-G, our proposed skeleton builder achieves a relative reduction of 32.7\% and 19.3\% in the RMSE indicator.\par

To further accentuate the superiority of our method, Figure \ref{fig.SB_show} provides a visual comparison between our method and other approaches on misspelled Chinese characters. Due to inherent bias in models and the high similarity between misspelled and right characters, existing methods such as RCN and RTN-G tent to automatically rectify misspelled characters to the right ones. As shown in the green boxes in Figure \ref{fig.SB_show}, RCN and RTN-G automatically correct glyphs that fall short of expectations. Moreover, when generating unseen glyph, RCN and RTN-G are unable to model all characters with high accuracy, resulting in distorted and tumultuous internal structures, as shown in the red boxes in Figure \ref{fig.SB_show}. In contrast, our proposed skeleton builder proves to be more stable and robust in modeling glyph imagery. The utilization of both radical and stroke level information mitigates the adverse effects of model bias.
\par

\subsection{Experiments on Font Generation}\label{sec.ex_fg}
In this section, we focus on evaluating the performance of our font generator on the font generation task, which aims to render Chinese characters in a standard font to other complex fonts.\par 

We first conduct experiments on the calligraphic font set, where the Chinese characters in the standard font are fed as inputs. Since our proposed font generator uses content features produced by skeleton builder, we utilize a DenseNet to extract content features from content images in order to compare fairly with other font generation methods.\par

\begin{table*}[!t]
	\caption{Performance comparison of state-of-the-art methods and our method on font generation task for seen characters.\label{tab.fg_seen}}
	\centering
	\begin{tabular}{c|c|cccc}
		\hline
		~ & \multirow{2}*{Methods} & \multicolumn{4}{c}{Seen Style}\\
		~ & ~ & RMSE $\downarrow$ & SSIM $\uparrow$ & LPIPS $\downarrow$ & FID $\downarrow$ \\ 
		\hline
		\multirow{8}*{\makecell{Font\\Generation}} 
          & Zi2zi \cite{zi2zi}      & 0.0255 & 0.7599 & 0.3125 & 68.24\\
		~ & 8-LF-Font \cite{LFfont} & 0.0246 & 0.7604 & 0.2722 & 57.71\\
		~ & EMD \cite{EMD}          & 0.0242 & 0.7610 & 0.2719 & 57.67 \\
		~ & DG-Font \cite{DGFont}   & 0.0236 & 0.7621 & 0.2471 & 32.15 \\
        ~ & FsFont \cite{FsFont} & 0.0227 & 0.7677 & 0.2228 & \textbf{7.31} \\
		~ & CG-GAN \cite{CGGAN}     & 0.0223 & 0.7696 & 0.2227 & 7.45  \\
        ~ & CF-Font\cite{CFFont}     & \textbf{0.0220} & 0.7704 & 0.2232 & 7.88 \\
		~ & \textbf{Font Generator}                   & 0.0233 & \textbf{0.7708} & \textbf{0.2224} & 7.33\\
  
		\hline
  
		\multirow{3}*{End-to-end} & RCN \cite{RCN} & 0.0281 & 0.7331 & 0.3235 & 64.69 \\
		~ & RTN-G \cite{RTN-G} & 0.0244 & 0.7608 & 0.2689 & 44.32 \\ 
		~ & \textbf{SFGN} & \textbf{0.0237} & \textbf{0.7631} & \textbf{0.2279} & \textbf{8.11} \\ 
		\hline
	\end{tabular}
\end{table*}

\begin{figure}[!t]
	\centering
	\includegraphics[width=0.85\linewidth]{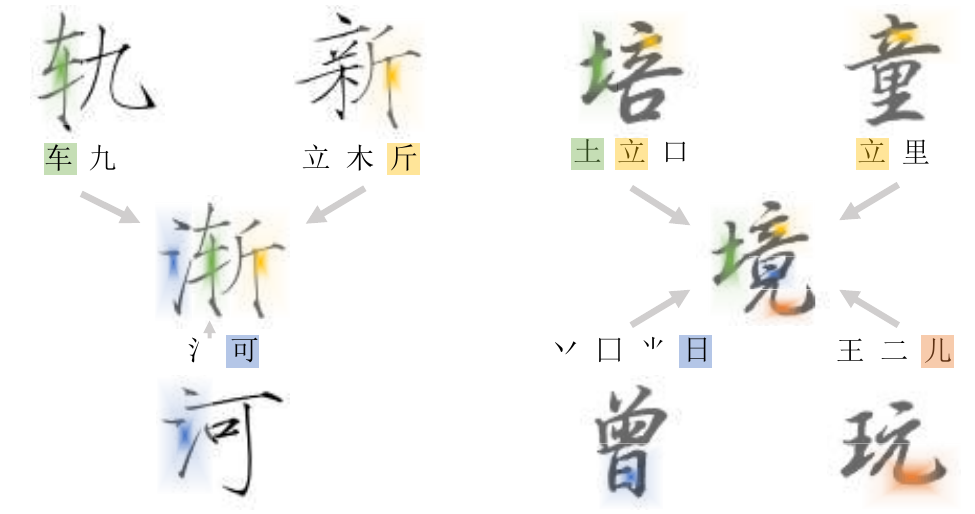}
	\caption{Visualization of the attention transmission path in the process of transitive attention mechanism.}
	\label{fig.Visual_TA}
\end{figure}

To comprehensively assess the effectiveness of our font generator, we compare it with other state-of-the-art methods, including arbitrary and single font generation methods. The quality of arbitrary generated images for seen and unseen characters is summarized in Table \ref{tab.fg_seen} and \ref{tab.fg_unseen}. We test both seen and unseen font styles and compare our font generator with state-of-the-art font generation methods. As shown in the top half of Table \ref{tab.fg_seen} and \ref{tab.fg_unseen}, the performance of proposed font generator is comparable to CG-GAN and significantly superior to other methods. Moreover, we denote the methods capable of achieving text-to-image font generation as end-to-end methods, such as RCN, RTN-G and our proposed SFGN. Our method emerges with the highest score among these methods, as evidenced by the bottom half of Table \ref{tab.fg_seen} and \ref{tab.fg_unseen}. Figure \ref{fig.Visual_TA} demonstrates the transmission path of attention during the process of the transitive attention mechanism. In contrast to the strategy adopted by FsFont, our method utilizes captions as the intermediary between content and style images. Richer informational cues and a precise attention transmission mechanism enable our method to achieve optimal performance. Notably, compared to other end-to-end methods, our proposed method can generate characters in unseen fonts. While SFGN may have lower performance than font generation methods, it does not rely on content image inputs. Our method has lower resource requirements and slight performance degradation which exhibits exceptional generation capabilities and few-shot learning proficiency. 

\begin{table*}[!t]
	\caption{Performance comparison of state-of-the-art methods and our method on font generation task for unseen characters.\label{tab.fg_unseen}}
	\centering
	\begin{tabular}{c|c|cccc}
		\hline
		~ & \multirow{2}*{Methods} & \multicolumn{4}{c}{Unseen Style} \\
		~ & ~ & RMSE $\downarrow$ & SSIM $\uparrow$ & LPIPS $\downarrow$ & FID $\downarrow$\\ 
		\hline
		\multirow{7}*{\makecell{Font\\Generation}} 
		~ & 8-LF-Font \cite{LFfont} & 0.0258 & 0.7588 & 0.2858 & 64.68\\
		~ & EMD \cite{EMD}          & 0.0252 & 0.7601 & 0.2612 & 64.84 \\
		~ & DG-Font \cite{DGFont}    & 0.0241 & 0.7612 & 0.2584 & 36.49 \\
        ~ & FsFont \cite{FsFont} & 0.0227 & 0.7620 & 0.2314 & 20.17\\
		~ & CG-GAN \cite{CGGAN}     & 0.0237 & 0.7618 & 0.2322 & \textbf{20.11}\\
        ~ & CF-Font\cite{CFFont}     & \textbf{0.0236} & 0.7633 & 0.2341 & 22.08\\
		~ & \textbf{Font Generator}                   & \textbf{0.0236} & \textbf{0.7637} & \textbf{0.2311} & 23.12  \\
  
		\hline
  
		End-to-end &  \textbf{SFGN} & \textbf{0.0248} & \textbf{0.7625} & \textbf{0.2389} & \textbf{28.22} \\ 
		\hline
	\end{tabular}
\end{table*}

\begin{table}[!t]
	\caption{RMSE comparison of state-of-the-art single font generation methods and our method.\label{tab.121}}
	\centering
	\begin{tabular}{c|ccccc}
		\hline
		Methods & ht$\rightarrow$ ls & ls$\rightarrow$ zk & zk$\rightarrow$ hp & hp$\rightarrow$ ht & avg \\ 
		\hline
        SGCE-Font \cite{SGCE} & 0.0137 & \textbf{0.0107} & 0.0196 & 0.0131 & 0.0143\\
        Strokegan \cite{strokegan} & 0.0242 & 0.0121 & 0.0344 & 0.0249 & 0.0239 \\
		\textbf{Font Generator} & \textbf{0.0117} & 0.0113 & \textbf{0.0156} & \textbf{0.0109} & \textbf{0.0124}\\
		\hline
	\end{tabular}
\end{table}

Additionally, Table \ref{tab.121} shows the performance on single font generation task. Four fonts selected from calligraphic font set are employed, including ``ht'' (HeiTi), ``ls'' (LiShu), ``zk'' (ZhengKai) and ``hp'' (HuPo). Our proposed method yields the highest scores in the majority of sub-tasks, and also demonstrates superior performance with the highest overall average score.\par

\begin{figure*}[!t]
	\centering
	\includegraphics[width=1.0\textwidth]{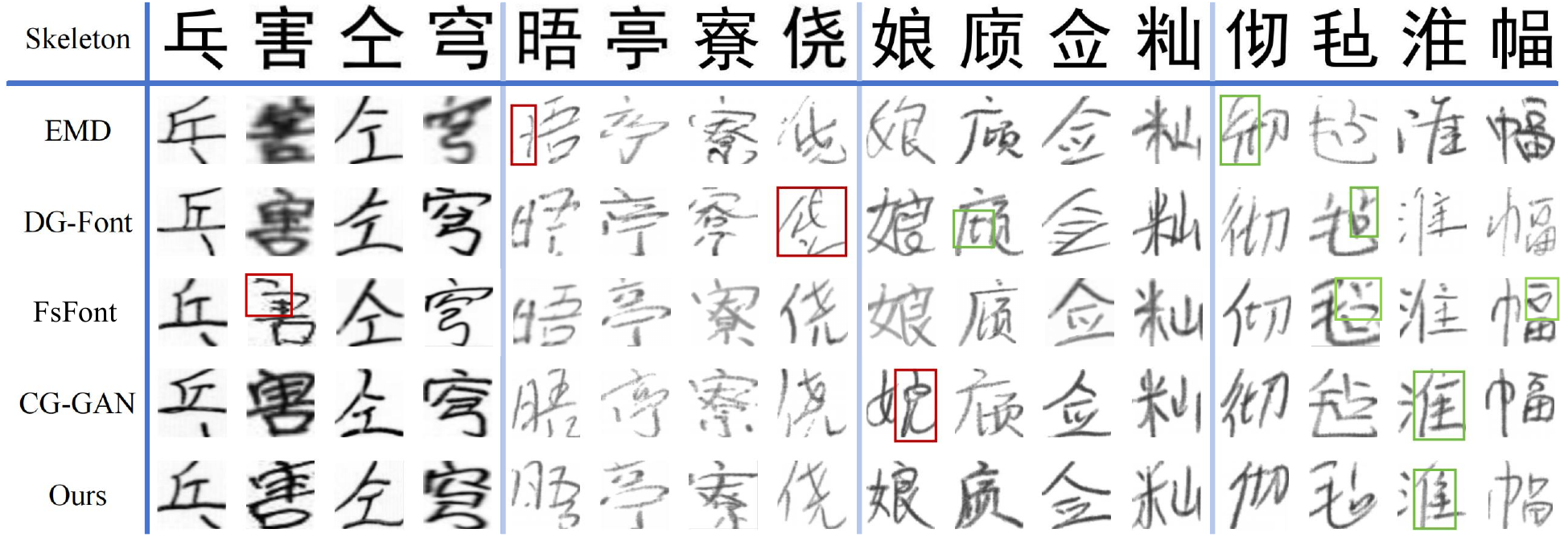}
	\caption{The generated handwritten characters of EMD \cite{EMD}, DG-Font \cite{DGFont}, FsFont \cite{FsFont}, CG-GAN \cite{CGGAN} and our proposed font generator. The results are separated into four groups with different writing styles. First three groups employ content images of unseen but existed Chinese characters and the fourth group employs the ones of misspelled characters. The content images are displayed on the top. The green boxes indicate cases where the font generation methods automatically correct the novel combinations of radical to the existed ones that appeared in training set. The red boxes indicate cases where the font transfer methods cannot effectively capture the glyph of content images.}
	\label{fig.FG_show_HW}
\end{figure*}

\begin{figure}[!t]
	\centering
	\includegraphics[width=0.6\linewidth]{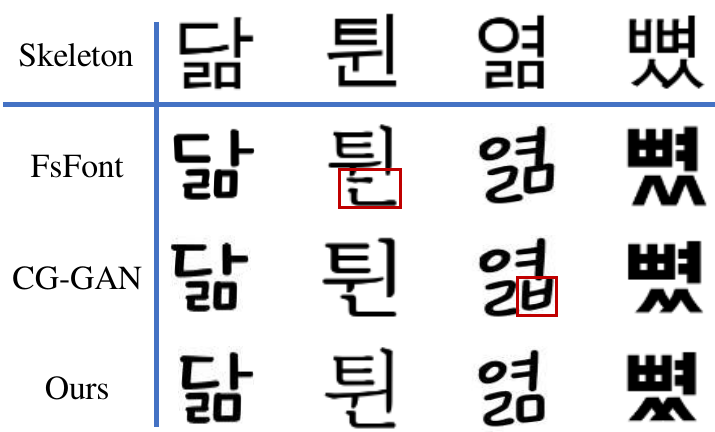}
	\caption{The generated unseen Korean characters of FsFont, CG-GAN and our proposed font generator. The red boxes indicate instances of incorrect generation.}
	\label{fig.FG_korean}
\end{figure}

Furthermore, we conduct an additional evaluation of font generation methods on misspelled characters, using the misspelled content images generated by our skeleton builder as input. The objective of this evaluation is to assess the ability of these methods to handle misspelled characters. Figure \ref{fig.FG_show} presents the generated results of various font transfer methods, where we utilize SFGN instead of a single font generator. Similar to the situation in glyph generation tasks, the existing font generation methods automatically correct misspelled characters that differ slightly from the right ones. As illustrated by the green box in Figure \ref{fig.FG_show}, the font generation models compensate for the missing strokes autonomously, which is attributed to the bias in the font generation models. Moreover, there are instances of structural confusion in the samples generated by font transfer methods, which results from a conflict between the input glyph and the model's internal representation. As depicted by the red box in Figure \ref{fig.FG_show}, other methods produce distorted outcomes with additional strokes. In contrast, our font generator is designed based on the concept of ``copying'',  which effectively averts these issues. Notably, even though FsFont also employs the ``copying'' methodology, its error rate is higher on misspelled characters. Due to the absence of constraints of radical-level captions, it tends to create seen categories from training set. The transitive-attention mechanism attenuates the negative impact of bias and makes our approach more robust for handling misspelled characters.\par


In addition, we show the performance of font generation methods on a handwriting dataset. Owing to the manifold diversity of handwritten characters, assessing the qualities of generated images using conventional metrics (RMSE, SSIM etc.) lacks persuasiveness. Figure \ref{fig.FG_show_HW} showcases the outcomes through different font generation methods. Compared with the experiments on calligraphic font set, the complexity of handwritten characters markedly attenuates the generalization capabilities of font generation models on handwritten scene. As discernible from the green boxes in Figure \ref{fig.FG_show_HW}, the font generation models rectify novel radical combinations to the ones that present in the training dataset. Especially, the four content images on the far right are the misspelled characters, where EMD, DG-Font, FsFont and CG-GAN all generate the error results that are the right character categories. Moreover, the red boxes in Figure \ref{fig.FG_show_HW} mark the distorted parts of the generated samples. Evidently, the proportion of characters that align with the intended content, generated by our SFGN, surpasses those of other methods. Nevertheless, despite SFGN's superior performance over other approaches, there still exists room for improvement in generating character with higher qualities and more accurate content.\par

Moreover, we conducted experiments on the Korean dataset to validate the generalization ability of our method across different languages. Figure \ref{fig.FG_korean} demonstrates the comparative performance of our proposed method against FsFont and CG-GAN, which are fine-tuned using Korean training set, based on the models trained on calligraphic font set. The quality of the generated unseen characters demonstrates the universality and accuracy of our approach in Korean.

\begin{table}[!t]
	\caption{The ablation experiment of the skeleton builder, comparing the impact of radical input ``$T_\text{R}$'', stroke input ``$T_\text{S}$'', bi-directional attention block ``BiDA'' and scale correction ``SC'' on the performance.\label{tab.ab_sb}}
	\centering
	\begin{tabular}{c|c|cc}
		\hline
		\multicolumn{2}{c|}{ } & RMSE $\downarrow$ & SSIM $\uparrow$\\ 
		\hline
        \multicolumn{2}{c|}{$T_\text{R}$} & 0.0176 & 0.8355 \\
        \multicolumn{2}{c|}{$T_\text{S}$} & 0.1325 & 0.2331 \\
        \hline
        \multirow{4}*{$T_\text{R}$ + $T_\text{S}$} & - & 0.0154 & 0.8576 \\
        ~ & BiDA & 0.0149 & 0.8723 \\
        ~ & SC & 0.0145 & 0.8817 \\
        ~ & SC + BiDA & \textbf{0.0142} & \textbf{0.8914} \\
		\hline
	\end{tabular}
\end{table}

\subsection{Ablation Study}\label{sec.ex_ab}
In order to ascertain the respective contributions of various components in both our skeleton generator and font generator, we conducted a series of ablation experiments to observe the performance of our model.\par

Our skeleton generator amalgamates both radical and stroke level information, resulting in more all-encompassing and intricate glyph representations. Moreover, our approach employs bi-directional attention block and scale correction to integrate stroke and radical information more smoothly. The experimental results are presented in Table \ref{tab.ab_sb}, where we denote radical input as ``$T_\text{R}$'', stroke input as ``$T_\text{S}$'', bi-directional attention block as ``BiDA'' and scale correction as ``SC''. When only radical inputs are used, the skeleton builder deteriorates into RTN-R \cite{RTN-R}. However, utilizing only strokes inputs results in poor performance, as stroke-level descriptions cannot uniquely represent each Chinese character. When radicals and stroke level information are combined, the RMSE index decreases to 0.0154, which effectively enhances the model's performance. Bi-directional attention block and scale correction also contribute constructively to information fusion, further reducing the RMSE 3.2\% and 5.8\% relatively. Overall, these ablation experiment results demonstrate that the integration of both radical and stroke level information significantly improves the performance of our skeleton generator.\par

\begin{figure}[!t]
	\centering
	\includegraphics[width=0.7\linewidth]{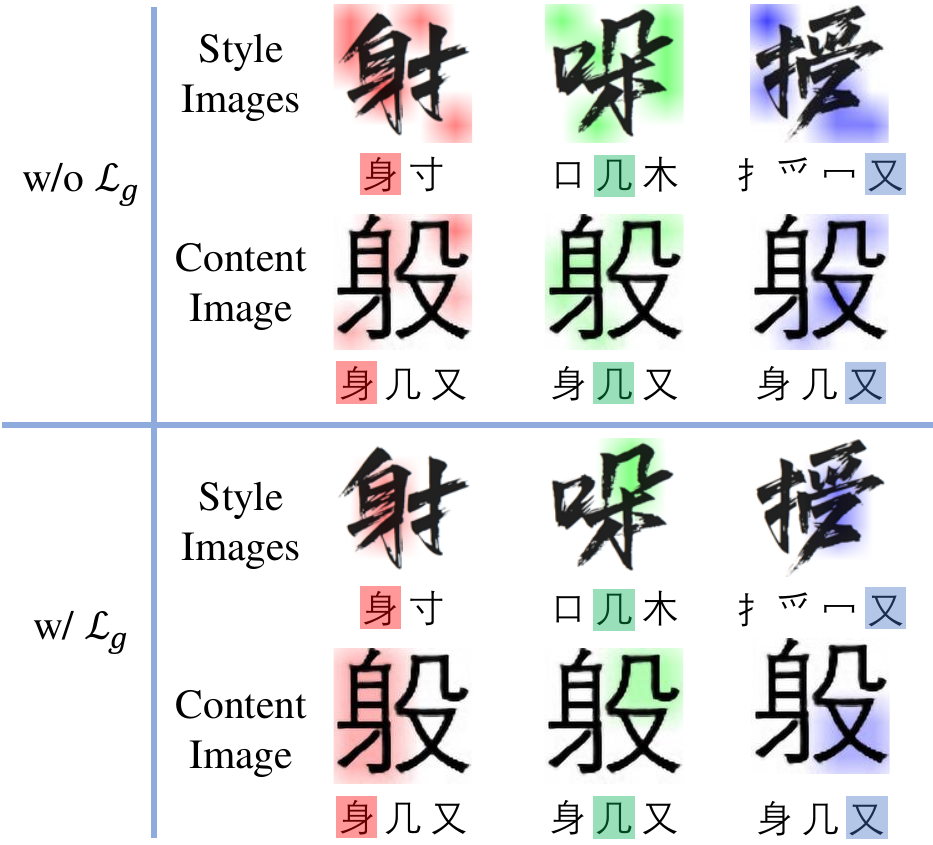}
	\caption{The difference in transitive-attention with or without using guided loss. Without guided loss, the attention relationships become very chaotic.}
	\label{fig.FG_attn}
\end{figure}

Three distinct objectives are employed during the training process: pixel loss $\mathcal{L}_p$, content loss $\mathcal{L}_c$ and guided loss $\mathcal{L}_g$. We conducted experiments to evaluate the model's performance in the absence of content and guided loss. The results of these experiments are shown in Table \ref{tab.ab_fg}. The utilization of content loss constrains the overall structure and optimizes the fine details of the generated Chinese characters, which results in a relative 4\% reduction in the RMSE of the skeleton builder. Guided loss is used to guide the optimization of the transitive-attention mechanism and align the components between the content images and style images. As illustrated in Table \ref{tab.ab_fg}, training the model without guided loss becomes highly unstable and challenging to achieve good performance. Similarly, as the attention mechanism of FsFont lacks the absence of supervision, its attention becomes diffused and capacity of capturing spatial relationship becomes worse in complex handwritten scene. In the Figure \ref{fig.FG_attn}, we visualize the transitive-attention, where the attention relationships become very chaotic without guided loss. Moreover, only with the assistance of guided loss, the model for font generation can be learned.\par 

\begin{table}[!t]
	\caption{Impact of pixel loss $\mathcal{L}_p$, content loss $\mathcal{L}_c$ and guided loss $\mathcal{L}_g$ on the performance of skeleton builder and font generator.\label{tab.ab_fg}}
	\centering
	\begin{tabular}{ccc|cc|cc}
		\hline
        \multirow{2}*{$\mathcal{L}_p$} & \multirow{2}*{$\mathcal{L}_c$} & \multirow{2}*{$\mathcal{L}_g$} & \multicolumn{2}{c|}{Skeleton Builder} & \multicolumn{2}{c}{Font Generator}\\
		~ & ~ & ~ & RMSE $\downarrow$ & SSIM $\uparrow$ & RMSE $\downarrow$ & SSIM $\uparrow$\\ 
		\hline
		$\checkmark$ & ~ & ~ & 0.0155 & 0.8572 & 0.2464 & 0.0331 \\
        $\checkmark$ & $\checkmark$  & ~ & \textbf{0.0142} & \textbf{0.8914} & 0.1912 & 0.1333 \\
        $\checkmark$ & ~ & $\checkmark$ & - & - & 0.0251 & 0.7595 \\
        $\checkmark$ & $\checkmark$ & $\checkmark$ & - & - & \textbf{0.0233} & \textbf{0.7708} \\
		\hline
	\end{tabular}
\end{table}

\subsection{Experiments on Chinese Character Error Correction}\label{sec.ex_ec}
Our proposed SFGN has proven to be successful in reducing bias in models and producing accurate generation of misspelled characters. In addition to the valuable contributions of font generation, we believe that misspelled characters have practical and educational value in the teaching of Chinese characters. In order to demonstrate the importance of misspelled characters and our method, we conduct a series of augmentation experiments on Chinese character error correction tasks, simulating the scene of students learning Chinese characters. An error correction model TAN \cite{TAN} is employed to verify such conclusion comprehensively. In the subsequent content of this section, we first introduce the augmentation data generated by SFGN. Then, we demonstrate the improved performance of the three sub-tasks, with the assistance of augmentation datasets.\par



Our proposed SFGN serves as the foundation for the generation of Chinese characters, which are subsequently utilized as data augmentation for recognition models. We divide Chinese characters into three distinct categories: misspelled characters, rare characters, and novel characters. Misspelled characters, which have minimal differences from their correct counterparts, are expected to yield the most notable improvement in the performance of the error correction task. Rare characters represent those that exist but have not been collected in the training set. Novel characters refer to those newborn characters that originate from the internet or are meaningless combinations of radicals. In total, we generated 20 writing styles for each of the three aforementioned categories, amounting to 11.4K samples and 570 distinct categories for the misspelled characters, approximately 288K samples and 14,430 categories for the rare characters, and approximately 582K samples and 29,101 categories for the novel characters.\par

\begin{table}[!t]
  \caption{Performance of TAN with the assistance of misspelled, rare and novel augmentation datasets on error correction task.\label{tab.subtask}}
  \centering
  \begin{tabular}{c|cc|ccc}
    \hline
    \multirow{2}*{\makecell{Augmentation\\Dataset}} & \multicolumn{2}{c|}{Right Set} & \multicolumn{3}{c}{Misspelled Set} \\
    ~ & $F_1$ & ACC & $F_1$ & ACC & CR\\ 
    \hline
    -          & 0.944 & 94.60 & 0.744 & 58.00 & 38.70 \\
    Misspelled & 0.945 & 94.97 & 0.762 & 63.30 & 39.92 \\
    Rare       & 0.946 & 95.00 & 0.764 & 63.41 & 40.07 \\
    Novel      & 0.947 & 95.09 & 0.766 & 64.52 & 40.75 \\
    Total      & \textbf{0.948} & \textbf{95.21} & \textbf{0.770} & \textbf{65.33} & \textbf{41.39}\\
    \hline
  \end{tabular}
\end{table}

As shown in Table \ref{tab.subtask}, our results demonstrate that the augmented data significantly improves the performance of TAN across all sub-tasks. Notably, the improvement in precision on the right testing set is relatively small, as the augmented data does not include any right character categories. The relative improvements in the $F_1$-score and accuracy rate on the right set are less than 1\%. Conversely, there is a substantial performance enhancement on the misspelled testing set, with a relative increase of at least 3.4\% in the $F_1$-score, 4\% in accuracy rate and 7\% in correction rate. Interestingly, the performance improvements of the 11.4K misspelled character dataset and 288K rare character dataset are similar, despite the latter being approximately 25 times larger in size. This can be attributed to the minuscule differences between misspelled and right characters, which improve the ability of TAN to distinguish similar radicals. Furthermore, novel set provides TAN with novel radical combinations in large quantities, also resulting in substantial performance improvements. By conducting these experiments, we simulate the impact of misspelled characters on students learning. Considering dataset capacity and training cost, misspelled characters are superior to other character categories in enhancing the recognition performance. The experimental conclusions also imply the educational value of misspelled characters.\par

Furthermore, we employed EMD, DG-Font, FsFont and CG-GAN to produce augmented datasets with misspelled characters, which aim to assess the impact of diverse generated datasets on error correction tasks. We anticipate that the generation errors depicted in the Figure \ref{fig.FG_show_HW} negatively influenced the performance of TAN, where error contents of generated samples causing more detrimental effect on performance. As illustrated in Table \ref{tab.augmentation}, the augmented dataset generated by SFGN exhibited the most significant assistance in the error correction tasks. Conversely, despite other methods exhibited excellent visual performance in generating tasks, they led to a degradation in the performance of TAN in sub-tasks.

\begin{table}[!t]
  \caption{Performance of TAN with the assistance of misspelled augmentation datasets generated by different font generation methods on error correction task.\label{tab.augmentation}}
  \centering
  \begin{tabular}{c|cc|ccc}
    \hline
    \multirow{2}*{\makecell{Augmentation\\Methods}} & \multicolumn{2}{c|}{Right Set} & \multicolumn{3}{c}{Misspelled Set} \\
    ~ & $F_1$ & ACC & $F_1$ & ACC & CR\\ 
    \hline
    -       & 0.944 & 94.60 & 0.744 & 58.00 & 38.70 \\
    EMD\cite{EMD}     & 0.926 & 94.44 & 0.738 & 60.17 & 38.73 \\
    DG-Font\cite{DGFont} & 0.920 & 94.03 & 0.731 & 61.22 & 38.84 \\
    FsFont \cite{FsFont} & 0.907 & 93.77 & 0.734 & 60.88 & 38.79 \\
    CG-GAN\cite{CGGAN}  & 0.944 & 94.58 & 0.745 & 61.51 & 39.00 \\
    \textbf{SFGN}    & \textbf{0.945} & \textbf{94.97} & \textbf{0.762} & \textbf{63.30} & \textbf{39.92} \\
    \hline
  \end{tabular}
\end{table}

\section{Conclusion and Future Work}
In this paper, we introduce a novel method named SFGN (Skeleton and Font Generation Network) for Chinese character generation. Our approach develops a skeleton builder to generate high-quality content features. Concurrently, a font generator is employed to achieve more robust font generation through a transitive-attention mechanism which facilitates the learning of alignment at radical-level between the content and style images. Our experimental results demonstrate that our proposed method outperforms state-of-the-art generation models on both glyph and font generation tasks. To further verify the educational and practical value of generated misspelled characters, we conduct a series of augmentation experiments on the handwritten Chinese character error correction tasks. The results of three sub-tasks validate the efficacy of misspelled characters, reflecting the merit of our proposed SFGN in Chinese character teaching.\par

At present, due to the complexity of handwritten characters, the performance of our proposed SFGN in handwritten scenes remains unsatisfactory. In the experiments, we showcase some results of failed modeling, which is also the research objective for the next phase. Additionally, effectively modeling Chinese characters on natural scene continues to be a challenging issue, owing to lighting, shadows, background, fonts etc. In the future research, we plan to develop a more effective Chinese character generation method to tackle the modeling of complex image attributes. Moreover, investigating the character generation on a line-level and reducing model costs constitute the next stage of research.




\bibliographystyle{elsarticle}
\bibliography{elsarticle}




\end{document}